\documentclass[runningheads,orivec]{llncs}
\usepackage{float}
\usepackage{placeins} 
\usepackage{multirow}
\usepackage[T1]{fontenc}
\usepackage{graphicx}
\usepackage{amsmath}
\usepackage{booktabs}
\usepackage{hyperref}
\usepackage{etoolbox}

 \let\subparagraph\paragraph
 \apptocmd{\thebibliography}{\scriptsize\setlength{\itemsep}{0pt}}{}{}
\usepackage[compact]{titlesec}
\usepackage{amsmath, amssymb} 
\usepackage{graphicx} 
\usepackage{algorithm} 
\usepackage{algpseudocode} 
\usepackage{algorithm}
\usepackage{algpseudocode}
\usepackage{amsmath} 
\usepackage{color}

\usepackage{wrapfig}

\usepackage{cite}

\begin{document}

\title{IQNN-CS: Interpretable Quantum Neural Network for Credit Scoring}
\titlerunning{IQNN-CS}
\authorrunning{A.S. Khan et al.}

\author{Abdul Samad Khan\inst{1} \and Nouhaila Innan\inst{2,3} \and Aeysha Khalique\inst{1} \and Muhammad Shafique\inst{2,3}}
\institute{ Lahore University of Management Sciences, Pakistan
\and 
eBRAIN Lab, Division of Engineering, New York University Abu Dhabi (NYUAD), Abu Dhabi, UAE\\ \and
Center for Quantum and Topological Systems (CQTS), NYUAD Research Institute, NYUAD, Abu Dhabi, UAE\\
\email{24120006@lums.edu.pk, nouhaila.innan@nyu.edu, aeysha.khalique@lums.edu.pk, muhammad.shafique@nyu.edu}}

\maketitle

\begin{abstract}
Credit scoring is a high-stakes task in financial services, where model decisions directly impact individuals' access to credit and are subject to strict regulatory scrutiny. While Quantum Machine Learning (QML) offers new computational capabilities, its black-box nature poses challenges for adoption in domains that demand transparency and trust. In this work, we present IQNN-CS, an interpretable quantum neural network framework designed for multiclass credit risk classification. The architecture combines a variational QNN with a suite of post-hoc explanation techniques tailored for structured data. To address the lack of structured interpretability in QML, we introduce Inter-Class Attribution Alignment (ICAA), a novel metric that quantifies attribution divergence across predicted classes, revealing how the model distinguishes between credit risk categories. Evaluated on two real-world credit datasets, IQNN-CS demonstrates stable training dynamics, competitive predictive performance, and enhanced interpretability. Our results highlight a practical path toward transparent and accountable QML models for financial decision-making.
\end{abstract}

\vspace{0.8cm}
\section{Introduction}
Credit scoring is a foundational task in financial services, directly influencing decisions related to loan approvals, credit limits, and interest rates \cite{thomas2017credit}. Its impact is not only financial but also societal, affecting individuals' access to economic opportunities. As such, credit scoring systems must meet two core requirements: high predictive performance and strict compliance with fairness and transparency regulations, such as the European Union's General Data Protection Regulation (GDPR) and the Basel III banking regulations. In this context, interpretability becomes essential, not as an optional enhancement, but as a mandatory feature for deployment in regulated environments.

Quantum Neural Networks (QNNs) have recently emerged as promising models for structured learning tasks, owing to their potential to encode and process high-dimensional data using fewer resources \cite{abbas2021power,innan2025next}. Their expressive power enables them to represent complex decision boundaries for domains like finance, where patterns in structured tabular data can be subtle and interdependent. 

However, existing QNN-based models often prioritize performance over interpretability, which is crucial for sensitive decision-making contexts such as credit scoring.
Moreover, the development of interpretable QNNs remains in its infancy. Unlike classical models, where attribution methods and explanation techniques are well-studied, the quantum learning community lacks standardized tools to reason about QNN decisions, especially in multiclass classification settings.
In this work, we ask a central question:
\begin{quote}
 \textit{How can QNNs be designed for real-world credit scoring applications, where achieving high accuracy is not sufficient, and interpretability is a core requirement?}
 \end{quote}
To address this, we propose \textbf{IQNN-CS}, an interpretable architecture of QNNs specifically tailored for multiclass credit scoring. Our goal is not to demonstrate quantum supremacy or to outperform classical machine learning models, but rather to explore how QNNs can be made suitable for high-stakes financial applications through interpretability-aware design.

\vspace{0.2cm}
\textbf{The main contributions of this paper are as follows:}
\begin{itemize}
\item We propose a hybrid classical–quantum pipeline where a variational QNN performs classification on structured financial data, with interpretability achieved using adapted classical and quantum techniques.
\item We introduce Inter-Class Attribution Alignment (ICAA), a new metric that quantifies attribution divergence across predicted classes, enabling structured analysis of model reasoning in multiclass tasks.
\item We evaluate IQNN-CS on two real-world credit datasets, focusing on robustness, attribution behavior, and interpretability, while emphasizing deployment feasibility in regulated financial settings.
\end{itemize}
\section{Background and Related Work}
\label{sec:background}
\subsection{QML for Credit Scoring}
Quantum Machine Learning (QML) has been increasingly applied in the financial sector \cite{biamonte2017quantum}, covering applications such as loan eligibility prediction \cite{innan2024lep}, credit default classification \cite{sengar2023comparative}, market prediction \cite{choudhary2025hqnn,pathak2024resource}, fraud detection \cite{innan2024financial1,innan2025qfnn,innan2024financial,alami2024comparative}, and other financial applications \cite{kashif2025evaluating, innan2025quantum}.
Some of these tasks are naturally formulated as classification problems, while others are formulated as combinatorial optimization problems. Within this broad spectrum, credit scoring has received limited but growing attention in QML research.

When it comes to credit scoring, only a few works have directly addressed the problem using quantum models. The systemic quantum score model introduced a quantum kernel-based approach aimed at improving generalization in low-data and imbalanced settings, demonstrating performance benefits in a production-grade financial scenario \cite{mancilla2024empowering}. A separate study explored a hybrid quantum–classical model for credit scoring among small- and medium-sized enterprises, highlighting reductions in training time by embedding a quantum layer into a classical neural network \cite{schetakis2024quantum}. Both approaches treated credit scoring as a binary classification task and focused on performance rather than transparency or explainability.

Other research efforts have framed credit scoring as an optimization problem. One study utilized a QUBO formulation to optimize scoring card thresholds and combinations, employing both quantum annealing and classical heuristics to enhance bank profitability under credit risk constraints \cite{zhang2024credit}. A related work investigated cost-effective feature selection for mobile credit scoring using a quantum-inspired evolutionary algorithm based on quantum gates and representations, achieving reduced feature costs and competitive accuracy \cite{chen2024quantum}. 

While these studies highlight the potential of quantum and quantum-inspired models for credit scoring, they either treat the task as binary classification or focus on optimization formulations. Crucially, they overlook the interpretability requirements critical for real-world deployment in regulated financial environments.
In contrast, we address credit scoring as a multiclass classification problem and focus on the interpretability of quantum models, an aspect that has not been covered in prior work. 
\subsection{Interpretability in QML}
Interpretability in QML is increasingly important for deploying models in sensitive domains such as finance. However, most classical explainability methods do not directly extend to quantum models due to their probabilistic nature and circuit-based representations.

Initial efforts, such as Q-LIME, adapt local explanation techniques to quantum settings by approximating the influence of features on predictions \cite{pira2024interpretability}. Other works combine quantum autoencoders with classifiers and apply LIME and SHAP for interpretability in quantum representation learning \cite{kottahachchi2025qrlaxai}. 

To address global explainability, frameworks like QuXAI use Q-MEDLEY to trace feature attributions through quantum feature maps \cite{barua2025quxai}. Visual tools such as QGrad-CAM further enable fine-grained class-specific explanations using variational quantum circuits \cite{lin2024quantum}.
Despite these advances, interpretability for structured, multiclass QML tasks, such as credit scoring, remains largely unexplored. Our work addresses this gap through a domain-specific, interpretable QNN framework.
\newpage
\section{Methodology}
\label{sec:methodology}

\begin{figure}[htbp]
    \centering
    \vspace{-0.7cm}
    \includegraphics[width=\textwidth]{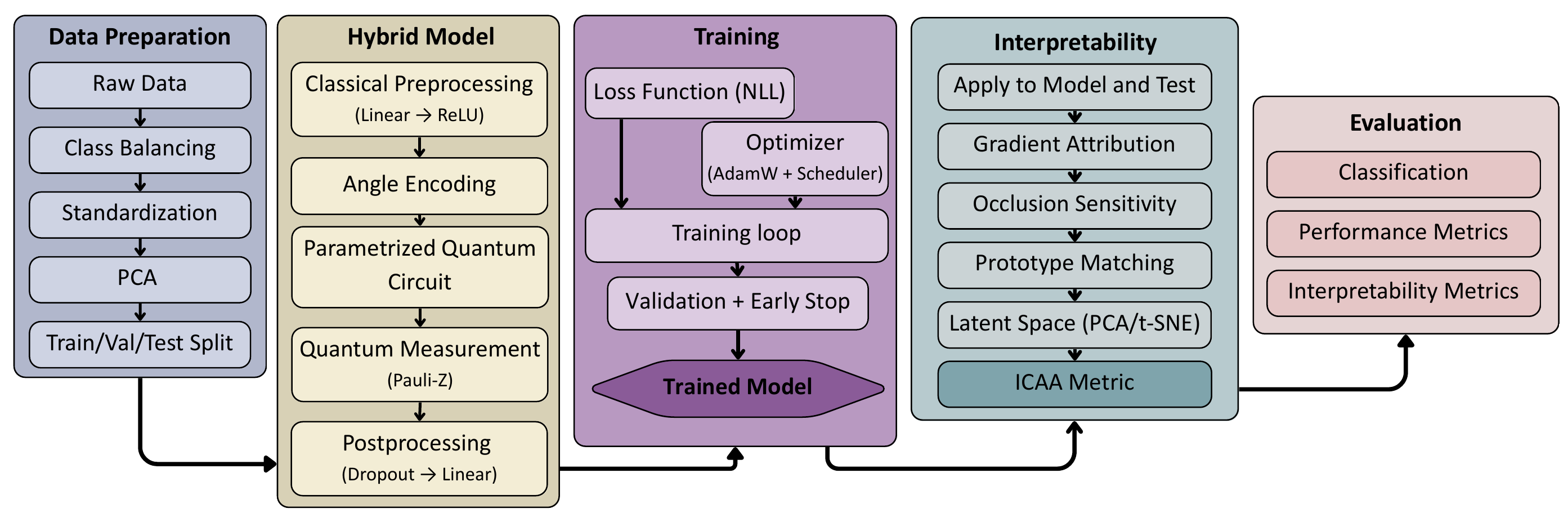}
    \vspace{-0.6cm}
    \caption{
         \scriptsize Overview of the IQNN-CS pipeline.}
             \vspace{-0.8cm}
    \label{fig:iqnn_methodology}
\end{figure}

The IQNN-CS framework is structured as a five-stage pipeline designed to deliver interpretable and accurate credit scoring using a hybrid quantum-classical architecture (see Fig. \ref{fig:iqnn_methodology} and Appendix Algorithm \ref{alg1}).
\subsection{Data Preparation}
The process begins with preprocessing two benchmark credit datasets containing numerical and categorical financial attributes. To address class imbalance, we apply undersampling to Dataset 1 and SMOTE to Dataset 2, depending on the severity of imbalance. All numerical features are standardized by transforming each feature $x_j$ to $x'_j = (x_j - \mu_j)/\sigma_j$, where $\mu_j$ and $\sigma_j$ denote the empirical mean and standard deviation. To reduce dimensionality and ensure compatibility with quantum hardware constraints, Principal Component Analysis (PCA) is applied, resulting in quantum-compatible input vectors. PCA projects the data $\mathbf{x}$ onto a lower-dimensional subspace using orthogonal components $\mathbf{w}_k$ as $z_k = \mathbf{w}_k^\top \mathbf{x}$, for $k = 1, \dots, d$. The resulting representation aligns with the available number of qubits. Stratified sampling is used to divide the dataset into training, validation, and test sets, preserving class distribution.

\subsection{Hybrid Quantum-Classical Model}
The model architecture consists of three sequential components. First, a classical preprocessing block transforms the PCA-compressed input $\mathbf{z} \in \mathbb{R}^d$ into an encoded latent vector $\mathbf{h}_{cl1} \in \mathbb{R}^{d'}$ through fully connected layers with ReLU activations. This output $\mathbf{h}_{cl1}$ is then passed to the quantum layer, where the features are encoded onto $N_Q$ qubits, typically initialized to the $|0\rangle^{\otimes N_Q}$ state, using angle encoding, where each input feature $\phi_j$ (from $\mathbf{h}_{cl1}$) rotates a qubit via an operation like $R_P(\phi_j) = e^{-i \phi_j P/2}$, with $P$ being a Pauli operator (e.g., $\sigma_x, \sigma_y, \sigma_z$). 
This is followed by a multi-layer entangling variational quantum circuit (VQC), $U(\boldsymbol{\theta})$, parameterized by trainable angles $\boldsymbol{\theta}$. Expectation values of Pauli-Z observables, $\langle \sigma_z^{(k)} \rangle = \langle \psi_0 | U^\dagger(\boldsymbol{\theta}) \sigma_z^{(k)} U(\boldsymbol{\theta}) | \psi_0 \rangle$ for each qubit $k$, are measured to extract quantum features. These features are concatenated and forwarded to the classical postprocessing head. This final block, which includes dropout regularization and linear projections, produces logits $\mathbf{o} \in \mathbb{R}^C$ over the $C$ credit risk classes. The full system is end-to-end differentiable and implemented via PennyLane's \texttt{TorchLayer}, enabling seamless integration with PyTorch's automatic differentiation capabilities.
\subsection{Training Procedure}
The model is trained using the negative log-likelihood (NLL) loss function, given for a batch of $M$ samples as
$L_{NLL} = -\frac{1}{M} \sum_{i=1}^M \sum_{c=1}^C y_{i,c} \log(\hat{y}_{i,c}),$ where $y_{i,c}$ is the ground-truth indicator for class $c$ and $\hat{y}_{i,c}$ is the predicted probability from the softmax output. Optimization is performed with AdamW, which introduces decoupled weight decay. Learning rate scheduling, such as cosine annealing, is employed to stabilize convergence.
Quantum gradients with respect to variational parameters $\boldsymbol{\theta}$ are computed using the parameter-shift rule:$\frac{\partial \langle \hat{O} \rangle}{\partial \theta_k} = \frac{1}{2s} \left( \langle \hat{O} \rangle_{\theta_k + s} - \langle \hat{O} \rangle_{\theta_k - s} \right),$ with $s = \pi/2$ for single-qubit gates. Early stopping is applied based on validation loss to prevent overfitting, and the best-performing model checkpoint is retained. 
\subsection{Post-Hoc Interpretability}
Interpretability is assessed using four complementary techniques (see Appendix Algorithm~\ref{alg2}). First, gradient-based attribution methods, including saliency maps, gradient$\times$input, integrated gradients, and SmoothGrad, estimate the importance of each input feature $x_j$ through $\partial L/\partial x_j$ or related formulations. 
Second, prototype matching is performed in the quantum feature space by extracting activation vectors $\mathbf{a}_{QNN}(x)$ and computing cosine similarity with training instances: $S_C(\mathbf{u}, \mathbf{v}) = \frac{\mathbf{u} \cdot \mathbf{v}}{\|\mathbf{u}\| \|\mathbf{v}\|}.$ This allows retrieval of similar historical examples for a given test input.  Fourth, we introduce the ICAA metric. For each class $c$, let $\mathbf{A}_c(x_{inst})$ be its attribution vector. The ICAA matrix is defined by: $ICAA_{ij}(x_{inst}) = S_C(\mathbf{A}_i(x_{inst}), \mathbf{A}_j(x_{inst})),$ capturing the pairwise similarity in feature attributions across classes. Additionally, we assess the region of indecision by perturbing inputs and measuring the variance in attribution vectors.
\subsection{Evaluation Metrics}
The final model performance is evaluated using accuracy and macro F1-score. Accuracy is computed as $
\text{Acc} = \frac{TP + TN}{TP + TN + FP + FN},$ where TP is the number of true positives, TN is the number of true negatives, FP is the number of false positives, and FN is the number of false negatives.
while the macro F1-score aggregates class-wise F1-scores: $F1_c = \frac{2 P_c R_c}{P_c + R_c}, \quad \text{Macro F1} = \frac{1}{C} \sum_{c=1}^C F1_c,$
where $P_c$ and $R_c$ are precision and recall for class $c$. Confusion matrices offer class-level diagnostic insights. Finally, we visualize interpretability outputs, gradient maps, and matched prototypes, along with latent space projections of QNN activations using t-SNE. These projections are color-coded by true labels to assess the separability of quantum features.
\section{Results}
\label{sec:results}

\subsection{Experimental Settings}
Experiments were conducted on two benchmark credit scoring datasets: Dataset 1 \cite{dataset1} and Dataset 2 \cite{dataset2}. All models were implemented using PyTorch and PennyLane \cite{bergholm2018pennylane}, with quantum circuits simulated via \texttt{default.qubit}. Training was executed on a CPU (Apple M3, 16 GB RAM) without GPU acceleration.
Each dataset was split into training (70\%), validation (15\%), and test (15\%) sets using stratified sampling. The hybrid model combined six qubits and four \texttt{StronglyEntanglingLayers} with classical dense layers and dropout. Optimization was performed over 50 epochs using the AdamW optimizer (learning rate 0.01) and class-weighted cross-entropy loss.
Full configuration details are provided in Appendix Table~\ref{tab:exp-settings}.

\subsection{Training and Convergence}
\label{ssec:training_convergence}
\begin{figure}[h]
    \centering
\includegraphics[width=1\textwidth]{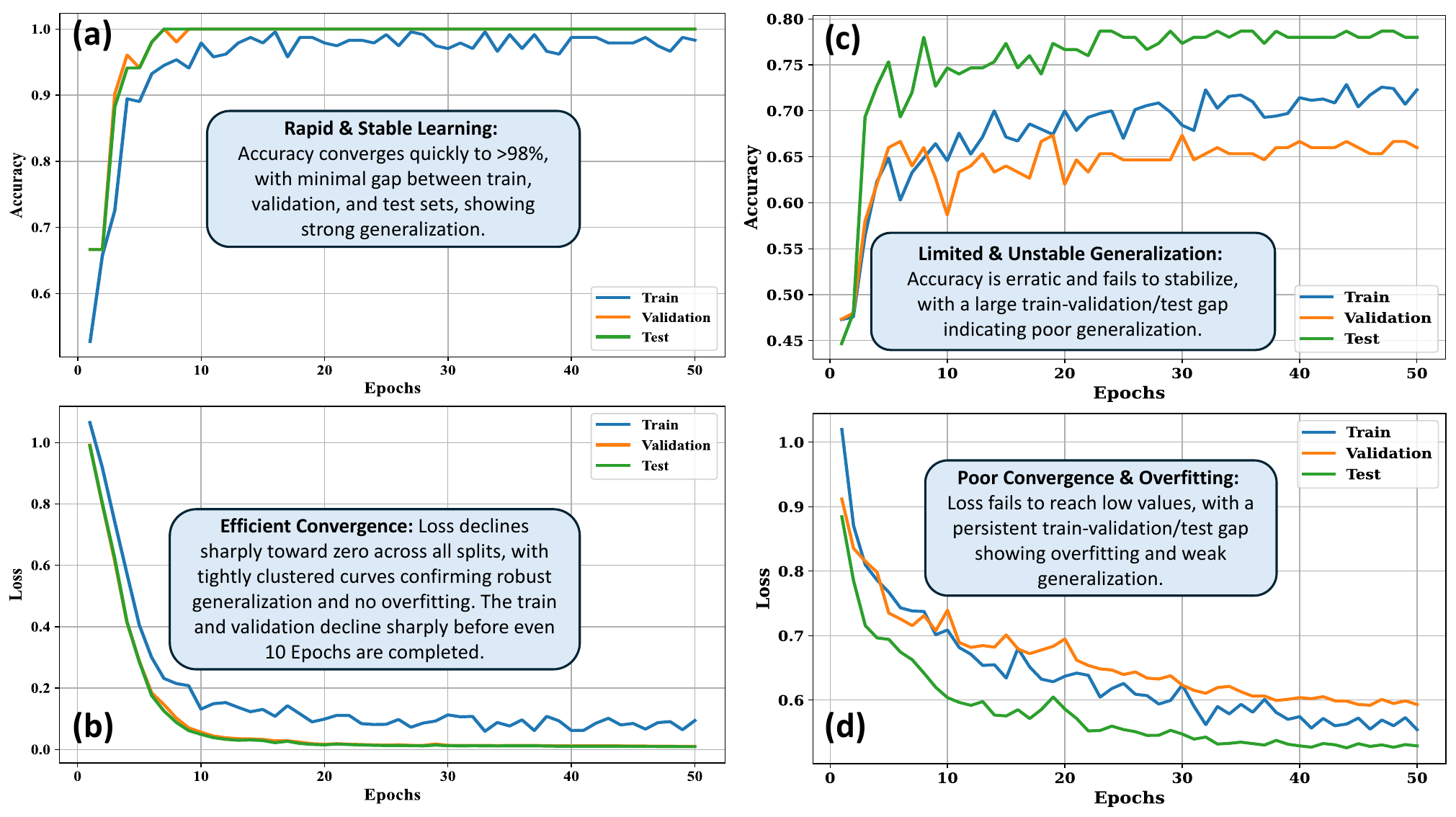}
\vspace{-0.8cm}
    \caption{\scriptsize Accuracy and loss curves for IQNN-CS. (a) accuracy and (b) loss for Dataset~1; (c) accuracy and (d) loss for Dataset~2.}
    \label{curv1}
\end{figure}

Fig.~\ref{curv1} shows the training, validation, and test curves for both datasets. For Dataset~1, accuracy curves in (a) exhibit rapid improvement within the first few epochs, stabilizing above 98\% across all splits. The corresponding loss curves in (b) show a sharp decrease followed by a smooth plateau, with minimal discrepancy between training and validation, indicating efficient convergence and strong generalization.
In contrast, Dataset 2 exhibits more unstable training behavior. Accuracy trends in (c) plateau early, with validation and test performance remaining below 80\%. The loss curves in (d) reveal a slower and noisier decline, particularly in the validation set. The wider gap between training and validation suggests either overfitting or greater dataset complexity due to feature diversity and class imbalance.

\subsection{Classification Performance}
Table~\ref{tab:classification_report} summarizes the classification results for both datasets. On Dataset~1, the model achieved perfect performance, with 100\% accuracy and F1-score. This confirms the model's ability to separate risk levels cleanly in well-structured data.
For Dataset~2, overall accuracy was 77.3\%, with variable class-wise metrics. The model showed high recall for the Low class (0.97) but low precision (0.64), suggesting misclassifications. The High class had strong precision (0.95) but lower recall (0.67), reflecting under-identification of high-risk cases. These results highlight Dataset~2's complexity and imbalance.
\vspace{-0.5cm}
\begin{table}[h]
\centering
\caption{\scriptsize Classification performance on both Datasets.}
\label{tab:classification_report}
\scriptsize
\begin{tabular}{lcccccc}
\toprule
\multirow{2}{*}{Class} & \multicolumn{3}{c}{Dataset~1} & \multicolumn{3}{c}{Dataset~2} \\
\cmidrule(lr){2-4} \cmidrule(lr){5-7}
 & Precision & Recall & F1-score & Precision & Recall & F1-score \\
\midrule
Low     & 1.00 & 1.00 & 1.00 & 0.64 & 0.97 & 0.77 \\
Average & 1.00 & 1.00 & 1.00 & 0.73 & 0.84 & 0.78 \\
High    & 1.00 & 1.00 & 1.00 & 0.95 & 0.67 & 0.79 \\
\midrule
\textbf{Accuracy (\%)} & \multicolumn{3}{c}{\textbf{100}} & \multicolumn{3}{c}{\textbf{77.3}} \\
\bottomrule
\end{tabular}
\end{table}
\vspace{-0.5cm}
\subsection{Interpretability Analysis}
To assess the transparency and trustworthiness of the IQNN-CS model, we conduct a comprehensive interpretability analysis. This includes examining latent quantum geometry, attribution consistency, example influence, class attribution alignment, and softmax entropy behavior. We compare findings across both datasets to understand how dataset quality impacts interpretability.

\begin{figure}[h]
    \centering
    \includegraphics[width=1\linewidth]{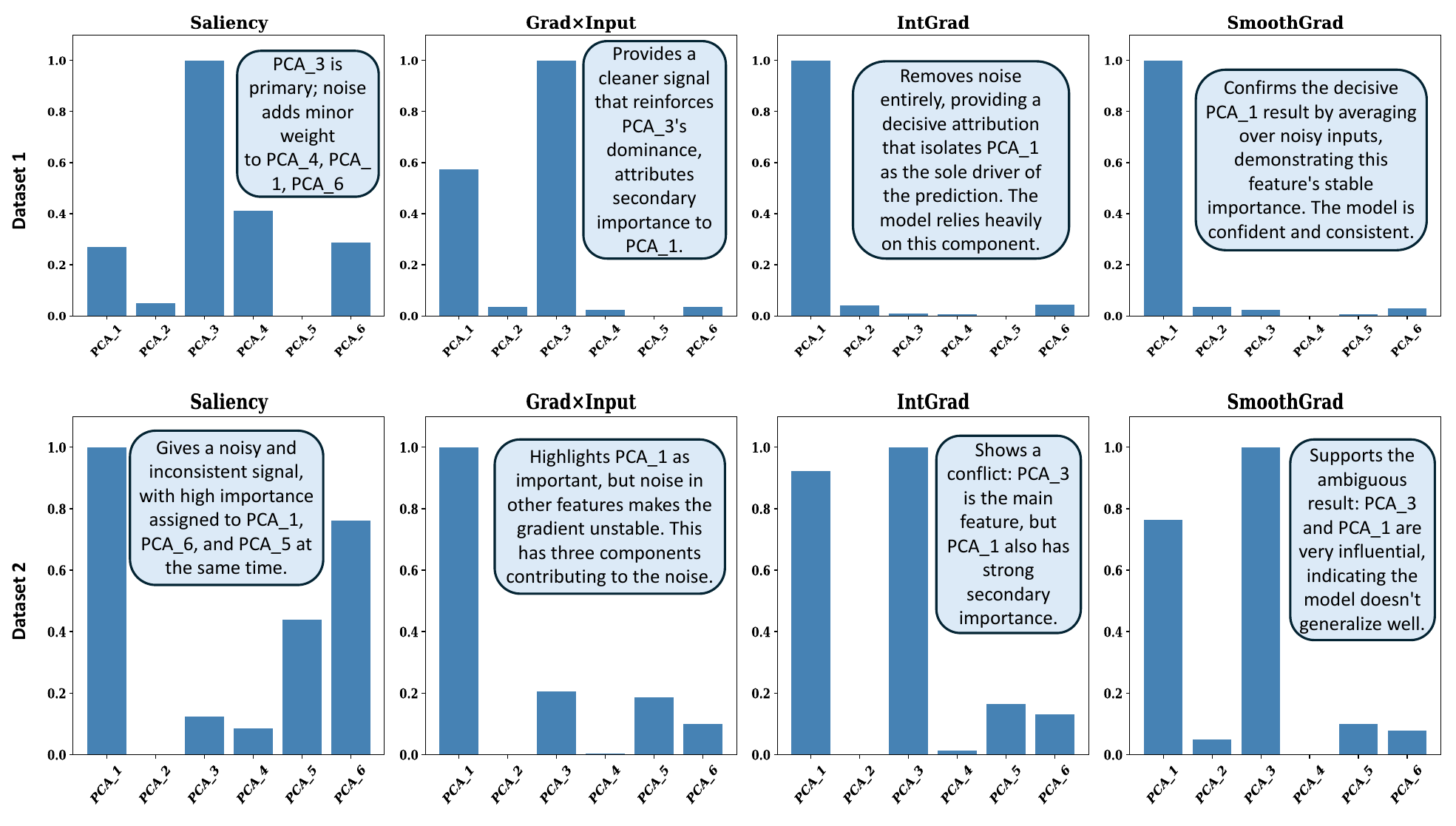}
    \vspace{-0.8cm}
    \caption{\scriptsize Saliency maps for a selected test sample. Dataset 1 shows concentrated attributions; Dataset 2 appears more diffuse.}
    \label{fig:sal}
\end{figure}

\noindent \textbf{Feature Attribution Analysis:}
As illustrated in Fig.~\ref{fig:sal}, saliency maps for a representative test instance show focused and interpretable patterns in Dataset 1. In contrast, Dataset 2 exhibits diffuse, noisy attributions that vary across runs, suggesting that the model relies less consistently on meaningful input features.
\vspace{-0.5cm}

\begin{figure}[h]
  \centering
  \vspace{-0.5cm}
  \includegraphics[width=1\linewidth]{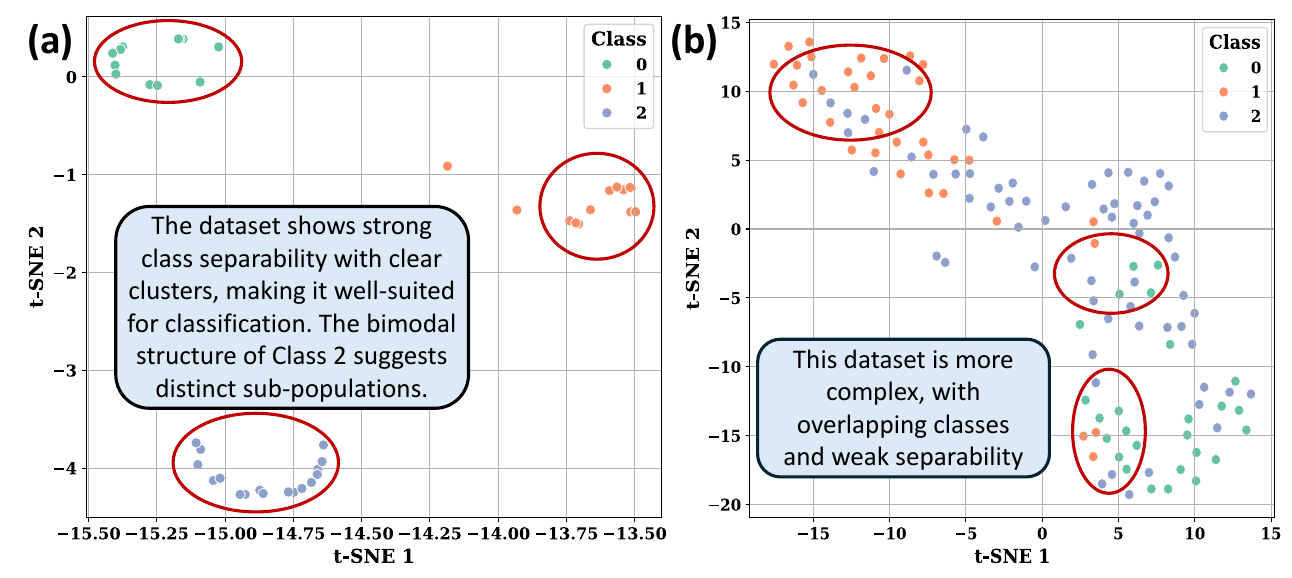}
  \vspace{-0.8cm}
  \caption{\scriptsize t-SNE projection of quantum embeddings. (a) Dataset 1 yields a clear separation; (b) Dataset 2 shows an overlap.}
  \label{fig:tsne}
\end{figure}

\noindent \textbf{Quantum Representation Geometry:}  
Latent embeddings extracted from the quantum layer are visualized using t-SNE as shown in Fig.~\ref{fig:tsne}. Dataset~1 shows well-separated manifolds, indicating robust internal representations. In contrast, Dataset~2 presents entangled clusters, particularly between the Average and High classes, reflecting the model's confusion in classification.

\begin{figure}[h]
  \centering
  \vspace{-0.5cm}
  \includegraphics[width=1\linewidth]{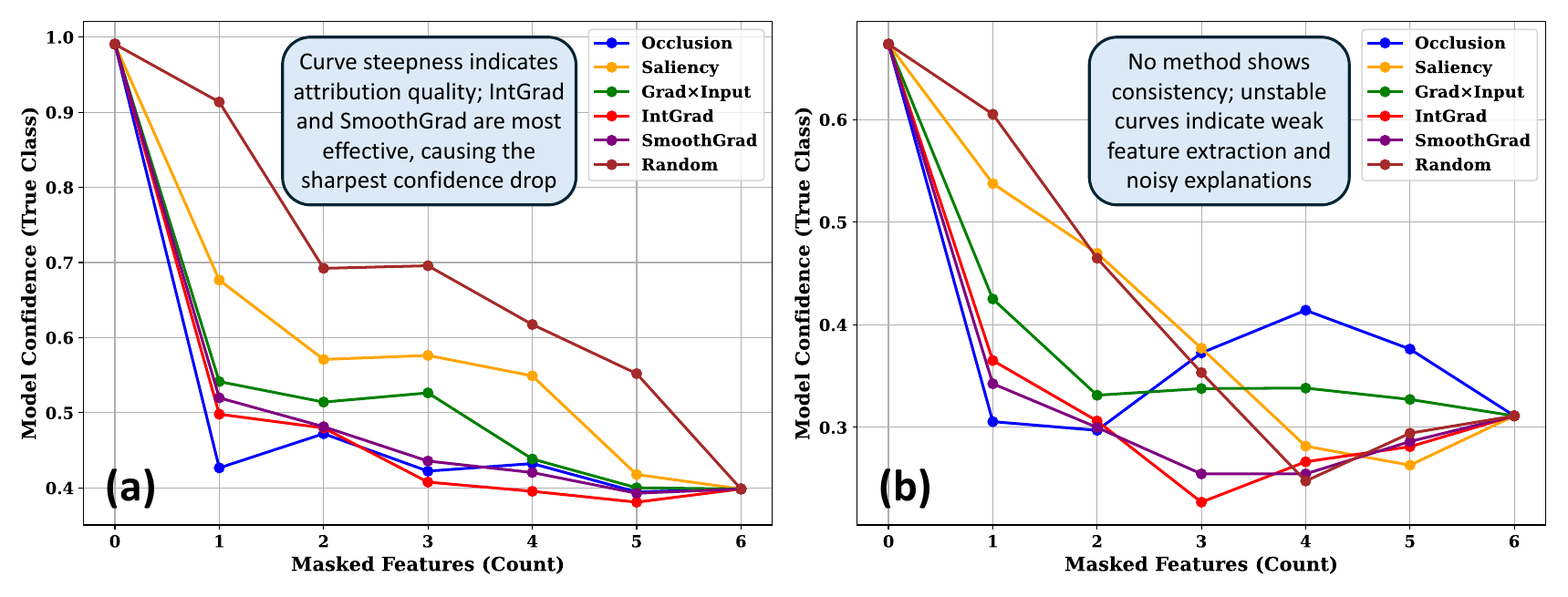}
  \vspace{-0.8cm}
  \caption{\scriptsize Prediction confidence vs. occluded features. Left: Dataset~1 shows sharp drops; Right: Dataset~2 decays smoothly.}
  \label{fig:prob_drop_combined}
\end{figure}

\noindent \textbf{Attribution Sensitivity:}  
To assess how strongly predictions depend on top-ranked features, we occlude inputs and track confidence degradation. As shown in Fig.~\ref{fig:prob_drop_combined}, Dataset~1 experiences sharp confidence drops, indicating reliance on a few informative features. In Dataset~2, degradation is gradual and less structured.

\begin{figure}[h]
  \centering
  \vspace{-0.5cm}
  \includegraphics[width=1\linewidth]{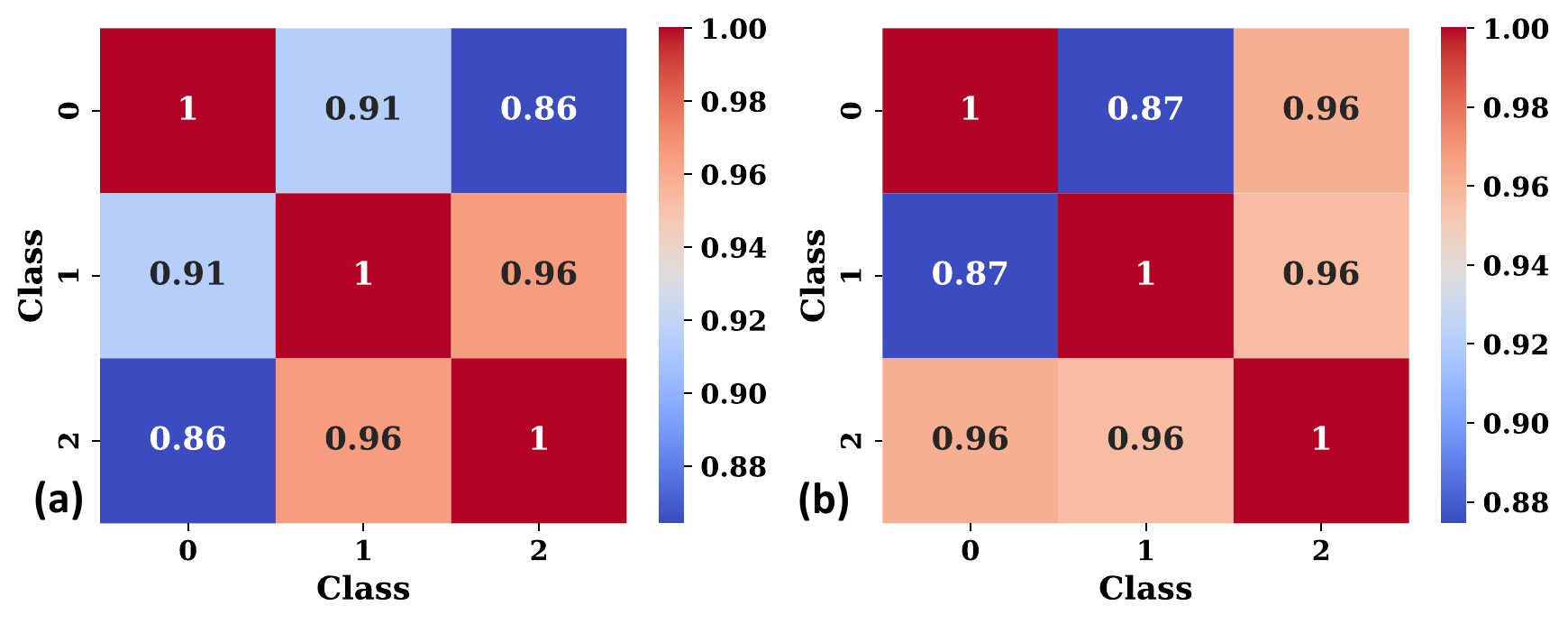}
  \vspace{-0.8cm}
  \caption{\scriptsize ICAA matrices. Dataset~1 (a) shows clean class-wise separation; Dataset~2 (b) reveals overlapping attribution logic.}
  \label{fig:icaa_combined}
\end{figure}

\noindent \textbf{ICAA:}  
We assess whether the model produces disentangled explanations for each class using ICAA. As shown in Fig.~\ref{fig:icaa_combined}, Dataset~1 yields low inter-class attribution similarity, while Dataset~2 exhibits significant overlap, revealing a less distinct decision rationale.

\subsection{Discussion}
Our experiments highlight a stark contrast in how the IQNN-CS model performs across two credit scoring datasets. While Dataset~2 proved more challenging, despite using the same model, its lower accuracy and less structured latent space suggest that the success of QNNs strongly depends on how well the data aligns with the model's inductive bias. Dataset~2 presented complex and overlapping feature spaces that the model struggled to disentangle.
Interpretability tools helped uncover these differences. Occlusion and our proposed ICAA metric revealed that, in Dataset~2, attribution patterns overlapped, indicating that the model relied on similar features across classes, a potential source of misclassification. Gradient-based explanations were inconsistent, especially in noisy predictions, suggesting they should be used cautiously in quantum settings. Additionally, example-based attribution showed that Dataset 2 often aligned with incorrect classes, despite high-confidence predictions. This shows that a model's internal logic can break down even when output probabilities appear reliable.

In summary, these findings underscore that interpretability is not simply about explanation; it is a crucial diagnostic tool. Metrics like ICAA provide deeper insight into how well a model separates reasoning across classes, which is particularly important in high-stakes domains like finance.

\section{Conclusion}
\label{sec:conclusion}
We introduced IQNN-CS, a hybrid quantum-classical model for credit scoring that combines strong predictive performance with interpretable outputs. Across two real-world datasets, the model demonstrated high accuracy when the data structure aligned with its quantum encoding and revealed clear decision patterns through interpretability analysis.
A key contribution of this work is the ICAA metric, which quantifies how distinctly the model reasons about different classes. It proved useful for identifying when the model's internal logic breaks down, even when standard metrics suggest good performance. Together with attribution sensitivity and example-based influence tracing, ICAA strengthens the tools available for understanding quantum models.
IQNN-CS shows that accurate quantum models can also be transparent and trustworthy. As quantum computing advances, such interpretability-first frameworks will be essential for deploying reliable AI in sensitive fields like finance and healthcare.

\section*{Acknowledgments}
This work was supported in part by the NYUAD Center for Quantum and Topological Systems (CQTS), funded by Tamkeen under the NYUAD Research Institute grant CG008, and the Center for Cyber Security (CCS), funded by Tamkeen under the NYUAD Research Institute Award G1104.

\bibliographystyle{ieeetr}
\bibliography{main}
\newpage
\appendix
\section*{Appendix}
\renewcommand{\thesubsection}{A\arabic{subsection}}
\appendix
\subsection{IQNN-CS Training Procedure}
The IQNN-CS architecture is trained using a hybrid quantum-classical pipeline as detailed in Algorithm~\ref{alg1}. 
\vspace{-0.7cm}
\begin{algorithm}
\scriptsize
\caption{IQNN-CS Training Procedure}\label{alg1}
\begin{algorithmic}[1] 
\Require Raw Datasets ($D_{raw1}, D_{raw2}$), NumEpochs ($N_{epochs}$), BatchSize ($B$), LearningRate ($\eta$), PCA Dimensions ($d_{PCA1}, d_{PCA2}$), Qubit Counts ($N_{Q1}, N_{Q2}$)
\Ensure Trained IQNN-CS model ($M_{IQNN-CS}$)
\State Initialize the IQNN-CS model structure ($M_{IQNN-CS}$):
\State \quad Define classical pre-processing network $M_{classical\_pre}$ using $d_{PCA}$ input dimensions.
\State \quad Define Quantum Neural Network $M_{QNN}$ with $N_Q$ qubits, specified layers, and encoding method.
\State \quad Define classical post-processing network $M_{classical\_post}$ for the specified number of output classes.
\State Assemble $M_{IQNN-CS}$ by sequentially connecting $M_{classical\_pre} \rightarrow M_{QNN} \rightarrow M_{classical\_post}$.
\ForAll{dataset $D_{raw}$ in $\{D_{raw1}, D_{raw2}\}$}
\State \Comment{Configure dataset-specific parameters: $d_{PCA}$, $N_Q$}
\State Perform preprocessing on $D_{raw}$ to obtain $D_{proc}$:
\State \quad Apply class balancing technique (e.g., Undersampling, SMOTE).
\State \quad Apply feature standardization (e.g., StandardScaler).
\State \quad Apply PCA to reduce dimensionality to $d_{PCA}$.
\State Split $D_{proc}$ into training $D_{train}$, validation $D_{val}$, and test $D_{test}$ sets.
\State Initialize optimizer (e.g., AdamW) with $M_{IQNN-CS}$ parameters and learning rate $\eta$.
\State Initialize learning rate scheduler (e.g., StepLR or CosineAnnealingLR).
\For{$epoch = 1 \to N_{epochs}$}
\State Set $M_{IQNN-CS}$ to training mode.
\ForAll{batch $(X_b, y_b)$ in $D_{train}$}
\State Clear gradients in the optimizer.
\State Obtain predictions $\hat{y}_b$ via a forward pass of $X_b$ through $M_{IQNN-CS}$.
\State Calculate loss (e.g., Negative Log-Likelihood) between $\hat{y}_b$ and $y_b$.
\State Compute gradients via hybrid backpropagation through $M_{IQNN-CS}$.
\State Update model parameters using the optimizer.
\EndFor
\State Adjust learning rate using the LR scheduler.
\State Set $M_{IQNN-CS}$ to evaluation mode.
\State Calculate validation loss $loss_{val}$ on $D_{val}$ using $M_{IQNN-CS}$.
\If{validation loss $loss_{val}$ meets early stopping criteria}
\State \textbf{break} \Comment{Cease training if validation performance degrades or stagnates}
\EndIf
\EndFor
\EndFor
\State \Return Trained $M_{IQNN-CS}$
\end{algorithmic}
\end{algorithm}
\vspace{-0.7cm}
\subsection{Interpretability Pipeline}
The interpretability module, summarized in Algorithm~\ref{alg2}, combines gradient-based saliency maps, occlusion analysis, example-based attribution, 
ICAA, and latent space visualization. These methods are executed on selected test instances, allowing insight into feature influence and consistency across representations.
\subsection{Experimental Settings}
The experimental configuration used across datasets is outlined in Table~\ref{tab:exp-settings}. This setup ensures consistency for reproducibility and fair interpretability comparisons.

\subsection{Extended Interpretability Analysis}

\subsubsection{Attribution Stability and Regions of Indecision:}
To assess robustness, we introduced Gaussian noise and computed the standard deviation of saliency maps. Most samples showed stable attribution under perturbation. However, Sample 12 in Dataset 2 showed significantly higher variance (Table~\ref{tab:region-indecision}), indicating unreliable class evidence.
\vspace{-0.55cm}
\begin{algorithm}[htpb]
\caption{Post-hoc Interpretability Analysis}\label{alg2}
\scriptsize
\begin{algorithmic}[1]
\Require Trained IQNN-CS model ($M_{IQNN-CS}$), Preprocessed Test Data ($D_{test}$), Set of test instances ($X_{interpret} \subseteq D_{test}$)
\Ensure Interpretability outputs (Attribution Maps, ICAA Matrix, Plots, etc.)
\State Set $M_{IQNN-CS}$ to evaluation mode.
\ForAll{test instance $x_{inst}$ in $X_{interpret}$}
\State \Comment{\textbf{1. Gradient-Based Attribution}}
\State For each target class $c$:
\State \quad Compute gradient-based attribution $A_c(x_{inst})$ for $x_{inst}$ towards class $c$ using $M_{IQNN-CS}$.
\State \quad Store or visualize $A_c(x_{inst})$ (e.g., as feature heatmaps).
\State \Comment{\textbf{2. Occlusion Analysis}}
\State Perform occlusion analysis on $x_{inst}$ with $M_{IQNN-CS}$ to obtain prediction probability drop curve $P_{drop\_curve}$.
\State \quad \Comment{This involves iteratively masking features of $x_{inst}$ and recording the drop in prediction probability.}
\State Store or visualize $P_{drop\_curve}$.
\State \Comment{\textbf{3. Example-Based Attribution (using QNN Activations)}}
\If{QNN activations for training data are not precomputed}
\State Extract QNN activations $Act_{train}$ from $M_{QNN}$ for all instances in $D_{train}$.
\EndIf
\State Extract QNN activation $act_{inst}$ from $M_{QNN}$ for instance $x_{inst}$.
\State Calculate cosine similarity scores $Sim_{scores}$ between $act_{inst}$ and all activations in $Act_{train}$.
\State Identify influential training examples based on $Sim_{scores}$.
\State \Comment{\textbf{4. Inter-Class Attribution Alignment (ICAA)}}
\State Let $A_0(x_{inst}), A_1(x_{inst}), \dots$ be the attribution vectors for $x_{inst}$ for each class (obtained from gradient-based attribution).
\State Calculate $ICAA_{ij} \gets \text{CosineSimilarity}(A_i(x_{inst}), A_j(x_{inst}))$ for all pairs of classes $(i, j)$.
\State Store the resulting $ICAA_{matrix}$ for $x_{inst}$.
\EndFor
\State \Comment{\textbf{5. Latent Space Geometry Visualization (on a subset of $D_{test}$)}}
\State Select a subset $X_{subset}$ from $D_{test}$.
\State Extract QNN activations $Act_{QNN\_subset}$ from $M_{QNN}$ for all instances in $X_{subset}$.
\State Apply a dimensionality reduction technique (e.g., t-SNE or PCA) to $Act_{QNN\_subset}$.
\State Plot the reduced-dimension embeddings, coloring them by true class labels.
\State \Return All generated interpretability maps, matrices, plots, and insights.
\end{algorithmic}
\end{algorithm}
\vspace{-1.41cm}
\begin{table}[H]
\centering
\scriptsize
\caption{\scriptsize Experimental Configuration for IQNN-CS}
\label{tab:exp-settings}
\vspace{-0.2cm}
\begin{tabular}{ll}
\toprule
\textbf{Component} & \textbf{Setting} \\
\midrule
Quantum Framework & PennyLane (default.qubit) \\
Number of Qubits & 6 \\
Quantum Layers & 4 (StronglyEntanglingLayers) \\
Embedding Strategy & AngleEmbedding \\
Classical Layers & Linear $\rightarrow$ ReLU $\rightarrow$ Linear + Dropout \\
Preprocessing & StandardScaler, OneHotEncoder, SMOTE \\
Dimensionality Reduction & PCA (6 components) \\
Batch Size & 16 \\
Optimizer & AdamW \\
Learning Rate & 0.01 \\
Scheduler & StepLR \\
Loss Function & CrossEntropy (class-weighted) \\
Epochs & 50 \\
Validation Strategy & 70-15-15 train-val-test split (stratified) \\
Random Seed & 42 (NumPy + PyTorch) \\
\bottomrule
\end{tabular}
\end{table}
\vspace{-1.2cm}
\begin{table}[H]
\scriptsize
\centering
\caption{ \scriptsize Standard deviation of saliency maps under 20 random Gaussian perturbations for selected test samples.}
\label{tab:region-indecision}
\vspace{-0.5cm}
\begin{tabular}{lccc}
\toprule
Sample & Dataset 1 (std) & Dataset 2 (std) & Indecisive? \\
\midrule
3  & 0.0493 & 0.0651 & No \\
7  & 0.1570 & 0.0395 & No \\
12 & 0.1426 & \textbf{0.2797} &Yes \\
\bottomrule
\end{tabular}
\vspace{-0.4cm}
\end{table}
\subsubsection{Influence of Training Examples on Test Samples:}

We analyze the influence of training samples on test instance 7 using test-to-train cosine similarity. As shown in Table \ref{tab2222}, in Dataset~1, the most influential samples align with the test sample's true class, indicating coherent internal representations. In contrast, Dataset~2 shows high-similarity points from multiple classes, suggesting ambiguity in the learned decision boundary.

\begin{table}[h]
\vspace{-0.4cm}
\caption{\scriptsize Cosine similarity between test sample 7 and top influential training examples in both datasets. Dataset~1 shows high similarity to same-class examples, while Dataset~2 yields a mix of same and different-class influences, suggesting less coherent internal representations.}
\label{tab2222}
\scriptsize
\centering
\vspace{-0.2cm}
\begin{tabular}{cccccl}
\toprule
Dataset & Test Sample & Train Index & Label & Cosine Sim. & Observation \\
\midrule
1 & 7 & 146 & 2 & 0.9989 & Same class as test \\
1 & 7 & 181 & 2 & 0.9976 & Same class as test \\
1 & 7 & 160 & 2 & 0.9976 & Same class as test \\
1 & 7 & 22  & 2 & 0.9930 & Same class as test \\
1 & 7 & 94  & 2 & 0.9416 & Same class as test \\
\midrule
2 & 7 & 503 & 1 & 1.0000 & Different class \\
2 & 7 & 19  & 1 & 1.0000 & Different class \\
2 & 7 & 433 & 1 & 1.0000 & Different class \\
2 & 7 & 588 & 1 & 1.0000 & Different class \\
2 & 7 & 173 & 2 & 1.0000 & Same class as test \\
\bottomrule
\end{tabular}
\vspace{-0.7cm}
\end{table}
\subsubsection{Prediction Confidence and Attribution Consistency}
\begin{figure}[h]
\centering
\includegraphics[width=1\textwidth]{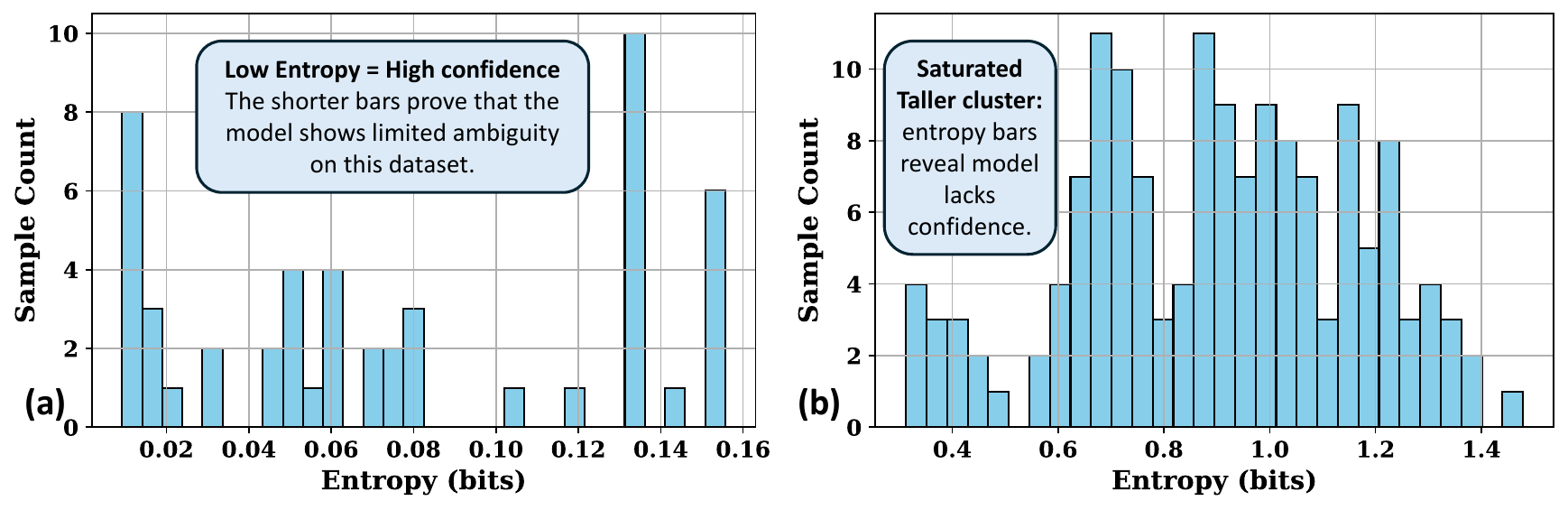}
\vspace{-0.76cm}
\caption{ \scriptsize Softmax entropy distributions. Dataset 1 (a) exhibits lower uncertainty compared to Dataset 2 (b).}
\label{fig:softmax}
\end{figure}

\begin{figure}[h]
\centering
\vspace{-0.7cm}
\includegraphics[width=1\textwidth]{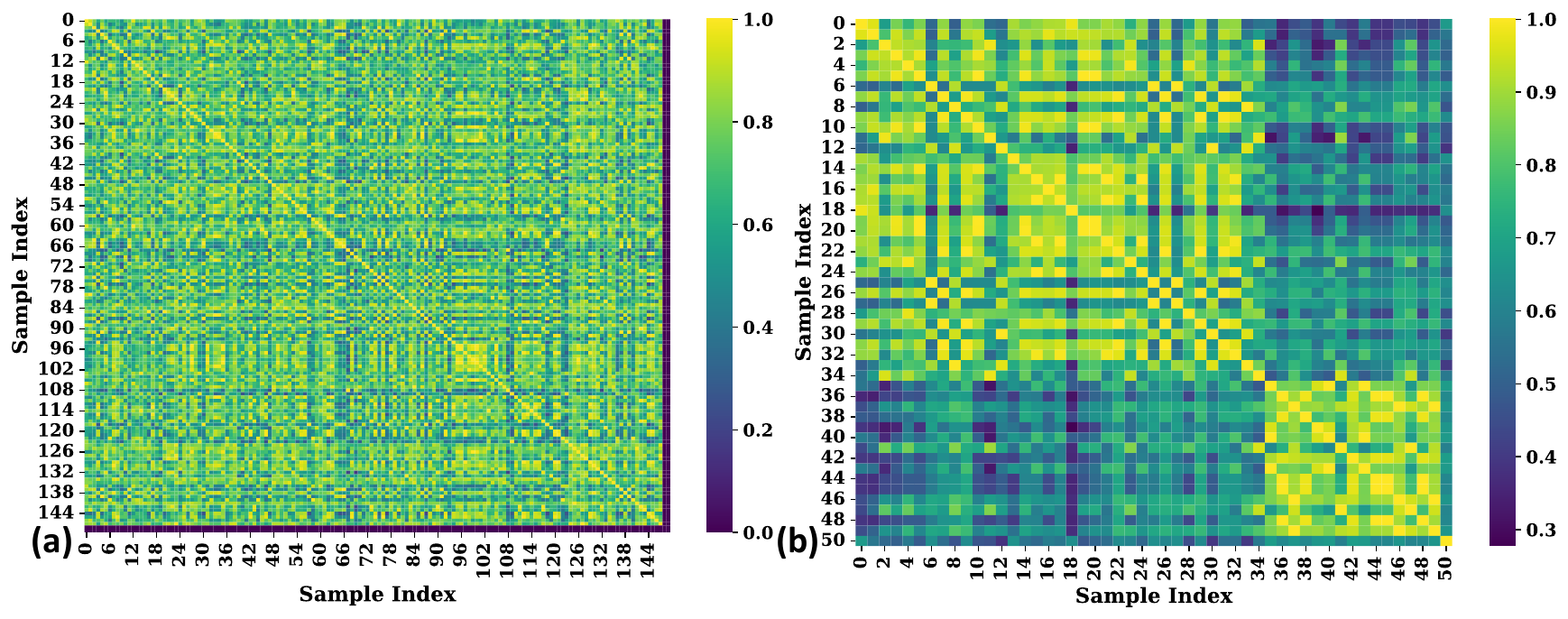}
\vspace{-0.7cm}
\caption{ \scriptsize Attribution similarity matrices across test samples for both Datasets: Dataset 1 (a) and Dataset 2 (b).}
\label{fig:similarity}
\end{figure}

Fig.~\ref{fig:softmax} shows that Dataset 1 had sharper softmax distributions, indicating high-confidence predictions. Dataset 2 yielded broader distributions, consistent with noisy convergence and less reliable model explanations. While the attribution similarity matrix (Fig.~\ref{fig:similarity}) reveals well-separated patterns in Dataset 1, while Dataset 2 lacks structure, underscoring inconsistency in explanation reliability.
\vspace{-0.6cm}
\vspace{-0.4cm}
\subsubsection{Method Evaluation Summary:}
Table~\ref{tab:interp_summary} synthesizes the comparative effectiveness of various interpretability techniques across both datasets. The evaluation is based on how well each method provides meaningful, class-consistent, and stable explanations aligned with model behavior.

In Dataset 1, most interpretability methods yielded coherent and high-confidence outputs. Occlusion and ICAA emerged as the most reliable techniques, offering sharply localized and class-discriminative explanations. Gradient-based methods such as saliency maps and integrated gradients also performed well, whereas SmoothGrad shows limited value due to noise sensitivity.
In Dataset 2, the utility of nearly all methods degraded, consistent with earlier findings of unstable training dynamics and less confident predictions. ICAA and occlusion retained partial utility but suffered from reduced clarity. Example-based attributions revealed inconsistencies in training-test relationships, and attribution stability flagged specific instances of model indecision.
Overall, the summary table highlights that interpretability reliability is closely tied to training convergence quality and dataset structure. Robust interpretability requires not only method design but also stable and semantically meaningful latent representations from the model itself.

\begin{table}[h]
\small
\centering
\vspace{-0.8cm}
\caption{ \scriptsize Interpretability method utility across both datasets.}
\vspace{-0.3cm}
\label{tab:interp_summary}
\scriptsize
\begin{tabular}{lcc}
\toprule
Method & Dataset 1 Utility & Dataset 2 Utility \\
\midrule
Saliency              & High                     & Medium \\
Gradient × Input      & Moderate                 & Low \\
Integrated Gradients  & High                     & Medium \\
SmoothGrad            & Low                      & Low \\
Occlusion             & Very High                & Medium \\
ICAA                  & Very High                & Medium \\
Example-Based         & High                     & Low \\
Indecision Detection  & Clean (all stable)       & Useful (flagged Sample 12) \\
\bottomrule
\end{tabular}
\end{table}
\newpage
\end{document}